%% file: ICRA_2025_Jonas.tex
\documentclass[letterpaper, 10 pt, conference]{ieeeconf}  

\IEEEoverridecommandlockouts                              

\usepackage{blindtext}
\usepackage{lipsum}  
\usepackage[nolist]{acronym}
\usepackage[table, dvipsnames]{xcolor}
\usepackage[pdftex]{graphicx} 
\usepackage{graphicx}
\usepackage[colorinlistoftodos]{todonotes}
\usepackage{graphicx}
\usepackage[caption=false,font=footnotesize]{subfig}

\graphicspath{{figures/}} 
\DeclareGraphicsExtensions{.pdf,.jpg,.png} 
\usepackage{amsmath} 
\usepackage{amsfonts} 
\usepackage{mathtools} 
\usepackage{tikz} 
\usetikzlibrary{shapes,arrows,fit,calc,positioning,automata,backgrounds}
\usepackage{color} 
\usepackage{multirow} 
\usepackage{hhline} 
\usepackage{booktabs}
\usepackage[bookmarks=false, breaklinks]{hyperref} 
\usepackage{enumerate} 
\usepackage{epstopdf} 
\usepackage{dsfont}
\usepackage{siunitx} 
\usepackage{cite}   
\usepackage{diagbox}    
\usepackage{pdfpages}   
\usepackage{breqn}
\usepackage{svg}
\usepackage{multicol}
\usepackage{gensymb}
\usepackage{siunitx}
\svgpath{{images/svg/}} 
\colorlet{Green1}{green!90!}
\colorlet{Green2}{green!60!}
\colorlet{Green3}{green!40!}
\colorlet{Green4}{green!20!}
\colorlet{Green5}{green!10!}

\usepackage{url}

\usepackage{breakurl}

\RequirePackage{tikz} 
\newcommand{\orcidicon}{\includegraphics[width=0.32cm]{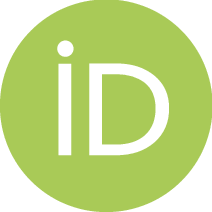}}

\foreach \x in {A, ..., Z}{%
\expandafter\xdef\csname orcid\x\endcsname{\noexpand\href{https://orcid.org/\csname orcidauthor\x\endcsname}{\noexpand\orcidicon}}
}

\definecolor{Bookcolor}{HTML}{00F9DE}
\definecolor{darkgreen}{rgb}{0.0, 0.75, 0.0}
\definecolor{gray}{gray}{0.9}
\definecolor{lightgray}{rgb}{0.86, 0.86, 0.86}

\makeatletter
\def\@citex[#1]#2{\leavevmode
\let\@citea\@empty
\@cite{\@for\@citeb:=#2\do
{\@citea\def\@citea{,\penalty\@m\ }%
\edef\@citeb{\expandafter\@firstofone\@citeb\@empty}%
\if@filesw\immediate\write\@auxout{\string\citation{\@citeb}}\fi
\@ifundefined{b@\@citeb}{\hbox{\reset@font\bfseries ?}%
\G@refundefinedtrue
\@latex@warning
{Citation `\@citeb' on page \thepage \space undefined}}%
{\@cite@ofmt{\csname b@\@citeb\endcsname}}}}{#1}}
\makeatother

\begin{acronym}
    \acro{GNN}{graph neural networks}
    \acro{CVAE}{convolutional variational autoencoder}
    \acro{GAT}{graph attention network}
    \acro{POMDP}{partially observable markov decision process}
    \acro{ICP}{iterative closest point}
    \acro{JCBB}{joint compatibility branch and bound}
    \acro{Conv1D}{one-dimensional convolutional}
    \acro{PPO}{proximal policy optimization}
    \acro{ORCA}{optimal reciprocal collision avoidance}
    \acro{NH-ORCA}{non-holonomic optimal reciprocal collision avoidance}
    \acro{DRL}{deep reinforcement learning}
    \acro{ICP}{iterative closest point}
    \acro{CBF}{control barrier function}
\end{acronym}
\begin{document}

%

\title{GIANT - Global Path Integration and Attentive Graph Networks for Multi-Agent  
Trajectory Planning}

\author{Jonas le Fevre Sejersen, Toyotaro Suzumura and Erdal Kayacan
\thanks{J. Fevre is with the Artificial Intelligence in Robotics Laboratory (AiR Lab), Department of Electrical and Computer          Engineering, Aarhus University,8000 Aarhus C, Denmark{\tt\small jonas.le.fevre at ece.au.dk} 
        T. Suzumura is with the Foundation Models for Artificial Intelligence group, Department of Information and Communication Engineering, Tokyo University, Tokyo, Japan. {\tt\small suzumura at ds.itc.u-tokyo.ac.jp}.
        E. Kayacan is with the Automatic Control Group, Department of Electrical Engineering and Information Technology, Paderborn University, Paderborn, Germany. {\tt\small erdal.kayacan at uni-paderborn.de}
}%
}
\maketitle
\begin{abstract}
\input{sections/abstract}
\end{abstract}
\IEEEpeerreviewmaketitle


\input{sections/introduction} 
\input{sections/related_work} 
\input{sections/method}
\input{sections/ablation_study}
\input{sections/experiments} 

\input{sections/results} 
\input{sections/conclusion}


\section*{Acknowledgment}
The authors would like to acknowledge the financial and hardware contribution from Beumer Group A/S, NVIDIA Corporation, and Innovation Fund Denmark (IFD) under File No. 1044-00007B.

\addtolength{\textheight}{-9cm}   
\bibliographystyle{IEEEtran}
\bibliography{References.bib}
%

\end{document}

%% file: sections/abstract.tex
This paper presents a novel approach to multi-robot collision avoidance that integrates global path planning with local navigation strategies, utilizing attentive graph neural networks to manage dynamic interactions among agents. We introduce a local navigation model that leverages pre-planned global paths, allowing robots to adhere to optimal routes while dynamically adjusting to environmental changes. The model’s robustness is enhanced through the introduction of noise during training, resulting in superior performance in complex, dynamic environments. Our approach is evaluated against established baselines, including NH-ORCA, DRL-NAV, and GA3C-CADRL, across various structurally diverse simulated scenarios. The results demonstrate that our model achieves consistently higher success rates, lower collision rates, and more efficient navigation, particularly in challenging scenarios where baseline models struggle. This work offers an advancement in multi-robot navigation, with implications for robust performance in complex, dynamic environments with varying degrees of complexity, such as those encountered in logistics, where adaptability is essential for accommodating unforeseen obstacles and unpredictable changes.

%% file: sections/introduction.tex

\section{Introduction}
\label{sec:introduction}

Collision avoidance is fundamental in autonomous robotics, especially in dynamic environments. Decentralized multi-agent reinforcement learning approaches have emerged as promising solutions. These approaches can be broadly categorized into end-to-end training from raw sensor data and using abstracted agent information. End-to-end approaches process raw sensor data, such as 2D laser scans or images, directly to produce actions \cite{DRL-Nav, DRL-NavV2, xie2023drl}. While this method efficiently handles both static and dynamic obstacles, it often lacks a higher-level interpretation of sensory data, impacting motion planning and interaction with the environment.

In contrast, agent-based approaches use abstracted information about agents and the environment, such as positions, velocities, and predicted trajectories~\cite{GA3C-CADRL, agent_based, wang2024multi, liu2023intention,le2024multi}. This higher-level representation aids decision-making processes, enabling better navigation and interaction with other agents. However, these methods may struggle with complex environments where abstraction might not capture all necessary details for optimal navigation.

Combining detailed static environment understanding from raw sensor data with high-level agent abstraction offers a promising approach for enhanced multi-agent navigation. This integration enables a comprehensive understanding of both static and dynamic environments. By representing agent intent through high-level abstraction, robotic fleets can achieve more effective cooperation, leading to improved navigation efficiency and coordination.


However, a critical challenge remains: ensuring efficient navigation over longer distances and in complex environments, which necessitates considering global path planning. Integrating global context to ensure local trajectories align with pre-planned global paths is crucial for optimizing navigation strategies, particularly in complex environments. Purely local navigation models, lacking global path information, are inherently susceptible to becoming trapped in local minima, especially within cluttered or dynamically changing scenarios. Consider a robot navigating a large warehouse to retrieve an item. Relying solely on local collision avoidance and a final goal location, it may become trapped in dead-end aisles or repeatedly circle obstacles, demonstrating suboptimal navigation due to the absence of global awareness~\cite{GP_as_goal}. In contrast, global planners effectively estimate time-efficient paths in multi-agent dynamic and noisy environments~\cite{le2023cameta,okumura2023lacam}. Therefore, the integration of global path planning with local navigation is not merely an incremental improvement, but a fundamental requirement for achieving truly robust and efficient navigation in realistic robotic applications.

\begin{figure}[t]
   \centering
   \includegraphics[width=0.5\textwidth]{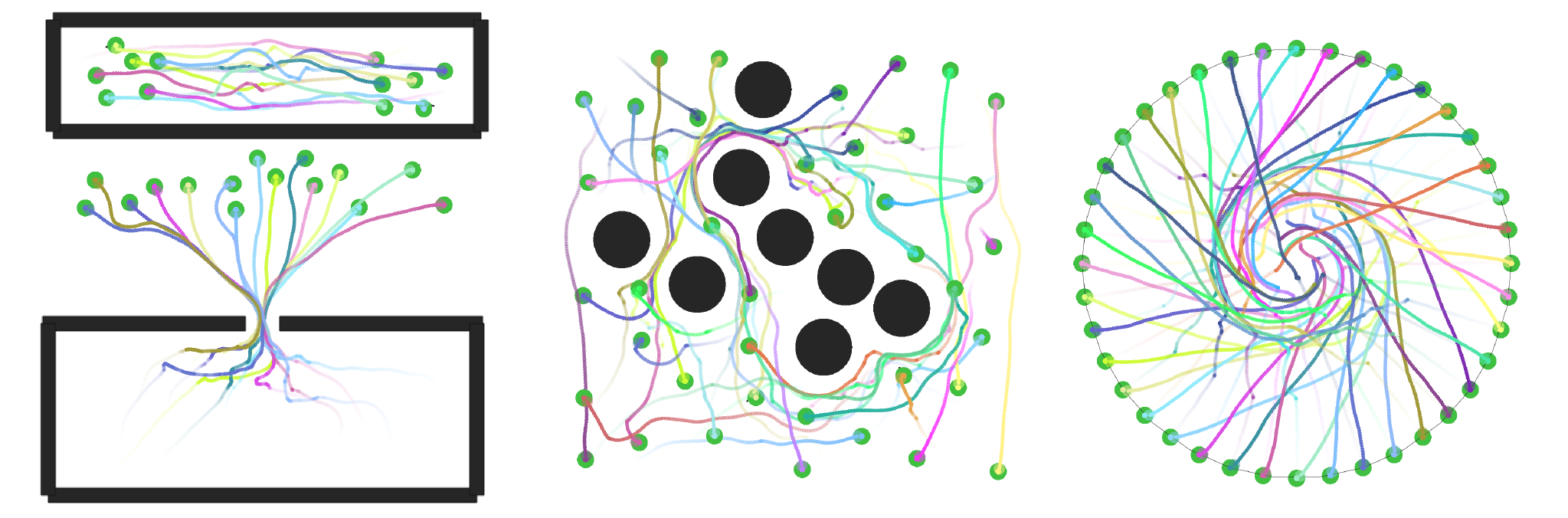}
   \caption{The figure illustrates the trajectory chosen by our model, where the robot navigates through highly dense and dynamic environments while adhering to the global path}
   \label{fig:overview}
   \vspace{-17pt}
\end{figure}


In this work, we introduce a local navigation model that combines global path planning with attentive graph neural networks to improve coordination and efficiency in multi-agent environments. As shown in Fig.~\ref{fig:overview}, our approach enables robots to navigate dense and dynamic environments while maintaining global path adherence and adapting to local obstacles. By leveraging graph-based representations, the model effectively captures interactions between agents, facilitating smoother and more cooperative multi-robot navigation.

\noindent The contributions of this study are the following:
\begin{itemize}
    \item \textbf{Novel local navigation model:} We introduce a learned local navigation model that incorporates preplanned global paths within the observation space, significantly enhancing the agent’s ability to navigate complex environments by maintaining adherence to planned routes while reacting to dynamic changes.
    \item \textbf{Graph structures and attentive neural networks:} Our model uses graph structures and attentive graph neural networks to manage an arbitrary number of neighboring agents, combined with LiDAR-based perception for complex obstacle avoidance.
     \item \textbf{Evaluation in structurally diverse simulations:}
     We rigorously evaluate our model across various structurally diverse simulated scenarios and noisy sensor information, demonstrating its superiority in safety, efficiency, and robustness.
    \item \textbf{Open-source implementation:} We provide an open-source implementation, supporting future research and benchmarking within the community.\footnote{\href{https://github.com/Zartris/MultiAgentRLCollisionAvoidance}{https://github.com/Zartris/MultiAgentRLCollisionAvoidance}} 
\end{itemize}

%% file: sections/related_work.tex
\section{Related work}
\label{sec:related_work}
The problem of multi-robot collision avoidance has been extensively studied, leading to the development of various strategies to address the challenges of dynamic environments and decentralized decision-making. Early approaches, such as \ac{ORCA}, employed heuristic methods to maintain safe trajectories, even in complex scenarios involving non-holonomic robots \cite{NH-ORCA, snape2010smooth, claes2012collision}. While effective, these methods often depend on precise perception systems and manually tuned parameters, limiting their scalability and adaptability in real-world applications.

\Ac{CBF} has recently gained attention in multi-agent systems for providing formal safety guarantees through analytically derived control laws~\cite{macbf_central,macbf_decentralized}. Beyond the classical reliance on precise system models, more recent research explores learning-based \ac{CBF}s that leverage machine learning to handle more complex dynamics~\cite{macbf_learning, harms2024neural}, and even uses graph neural networks to create neural graph \ac{CBF} for scalable, decentralized collision-avoidance~\cite{macbf_neuralbarriers}. Additionally, online \ac{CBF}s can adapt their hyperparameters in real-time, mitigating issues that arise from static, pre-tuned safety constraints~\cite{macbf_online, sweatland2024adaptive}. 
While these \ac{CBF}-based approaches offer valuable theoretical or empirical guarantees for collision avoidance, they often require precise system models or parameter tuning to remain effective in highly dynamic or uncertain conditions.

In contrast to the model-based nature of \ac{CBF}, model-free control offers an attractive alternative by directly learning multi-agent collision avoidance policies without explicit reliance on system models. \Ac{DRL} model-free controllers have demonstrated their effectiveness in handling complex navigation tasks, achieving real-time path planning from raw sensor data, and providing adaptability to dynamic obstacles \cite{DRL-Nav, DRL-NavV2, xie2023drl}. Notably, recent work has further demonstrated \ac{DRL}'s ability to navigate highly complex, real-world scenarios, such as pedestrian-rich environments, where interactions and decision-making around humans are crucial~\cite{everett2021collision, flogel2024socially}. However, the downside of learning-based controllers is that it is difficult to justify the actions due to its black box nature.

Beyond the challenges of navigating dynamic environments and interacting with complex agents, cooperative behaviors are crucial for efficient multi-robot navigation. To address this, \ac{GNN} have been employed to model interactions in multi-robot systems, significantly improving cooperative behaviors in navigation.
The work by~\cite{li2020graph, li2021magat} showcased the ability of \ac{GNN}s to distribute local information among robots, enabling more efficient local navigation. However, these methods heavily rely on communication between robots, which can be a limitation in scenarios where communication is not reliable or feasible. In contrast,~\cite{chen2020relational} proposed a relational graph learning approach for crowd navigation that operates without explicit communication between agents, inferring relationships based on observed states. While this approach improves efficiency in dynamic environments, it lacks environmental awareness, focusing primarily on agent interactions without adequately accounting for static obstacles. This limitation often leads to difficulties in navigating through environments with significant static structures.

While graph neural networks address local interactions, global path planning is essential for optimizing navigation efficiency in multi-robot systems~\cite{le2023cameta, chen2024traffic}. Many model-free local navigation methods, particularly \ac{DRL}-based approaches, often overlook global context, leading to suboptimal overall navigation. To mitigate this,~\cite{GP_as_goal} proposed utilizing a global planner to set intermediate goals for local models. While reducing local minima traps, this approach can lead to over-focus on temporary goals, potentially decreasing overall navigation efficiency.
The method proposed in this paper seeks to better integrate global planning with local decision-making, specifically within a model-free framework, aiming to address these limitations and enhance overall model-free navigation efficiency.

\section{Problem formulation}
\label{sec:problem_formulation}
This study addresses the challenge of non-communicating, multi-robot collision avoidance for nonholonomic differential drive robots navigating a continuous Euclidean plane populated with static obstacles and other decision-making robots. Each robot is modeled as a homogeneous disc with a uniform radius \( R \). At each timestep \( t \), the \( i \)-th robot receives an observation \( o_t^i \) and computes a collision-free steering command \( a_t^i \) to move towards its goal \( g_i \) from its current position \( p_t^i \). The observation \( o_t^i \) is a probabilistic representation of the system state \( s_t^i \), providing partial information due to the lack of explicit knowledge about the states and intentions of other robots.

The multi-robot collision avoidance problem is formulated as a sequential \ac{POMDP} characterized by the tuple \( (S, A, P, R, \Omega, O) \). The state space \( S \) includes all possible configurations of robots and obstacles. Each state \( s_t^i \) encompasses the robot’s position, velocity, orientation, and the positions and velocities of surrounding robots and static obstacles. The action space \( A \) comprises translational and rotational velocities \( [v_t^i, \omega_t^i] \). Given a partial observation \( o_t^i \), each robot independently computes an action \( a_t^i \) sampled from a shared stochastic policy \( \pi \):
\begin{equation}
    a_t^i \sim \pi_\theta(a_t^i | o_t^i)
\end{equation}
where \( \theta \) denotes the policy parameters. The state transition function \( P(s_{t+1}^i | s_t^i, a_t^i) \) captures the probabilistic evolution of the state based on the chosen action and environmental dynamics. The reward function \( R \) balances goal achievement, collision avoidance, social distance, and progress along the path:
\begin{equation}
    r(s_t^i, a_t^i) = r_{\text{goal}} + r_{\text{collision}} + r_{\text{social}} + r_{\text{progress}}
\end{equation}

The observation space \( \Omega \) models the partial observability due to sensor noise and occlusions. The objective is to find an optimal shared policy \( \pi^* \) that maximizes the expected mean reward for all robots:
\begin{equation}
    \pi^* = \arg\max_\pi \mathbb{E} \left[ \sum_{t=0}^\infty \frac{1}{N} \sum_{i=1}^N \gamma^t r(s_t^i, a_t^i) \right]
\end{equation}
where \( \gamma \) is the discount factor. This \ac{POMDP} framework effectively captures the complexity and uncertainty inherent in dynamic, multi-agent environments. It integrates global and local contextual information, adeptly handles trajectory uncertainties, and fosters cooperative behaviors among robots.

%% file: sections/method.tex
\section{Methodology}
\label{sec:methodology}
Building upon the motivation outlined in Section \ref{sec:introduction} for integrating global context into local navigation, our methodology is structured as a two-stage approach. First, to provide high-level guidance and ensure efficient navigation in complex environments, we address global path planning. Given that our robots operate without communication, we utilize standard, $A^*$ search for global path planning for each robot, considering only the static environment map and ignoring the presence of other agents. These independently planned global paths serve as a baseline trajectory for each robot, guiding them towards their goals while avoiding known static objects.

Building upon this global path planning stage, the core of our method lies in a decentralized local navigation model, trained using reinforcement learning. This local model is responsible for dynamic collision avoidance and navigation in the presence of other agents and unmapped obstacles, using a rich observation space to perceive the immediate environment and make reactive control decisions.  The subsequent sections detail the design of this observation space, the action space, the neural network architecture, and the reinforcement learning training process.
\subsection{Observation space}
\begin{figure}[!b]
    \centering
    \vspace{-4mm}
    \subfloat[]{%
        \includegraphics[width=0.30\linewidth]{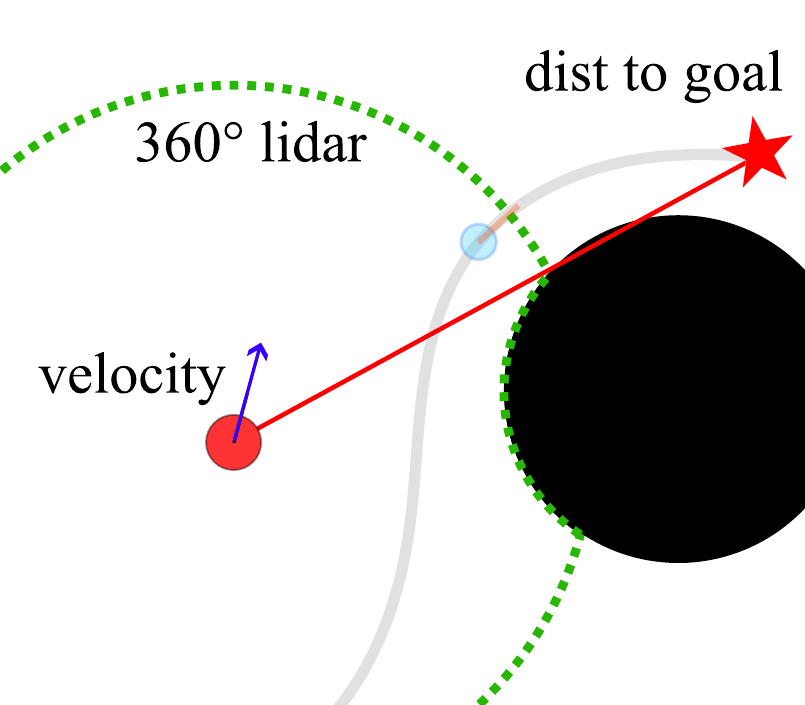}%
        \label{fig:obersvations-lidar}%
    }
    \hfill
    \subfloat[]{%
        \includegraphics[width=0.30\linewidth]{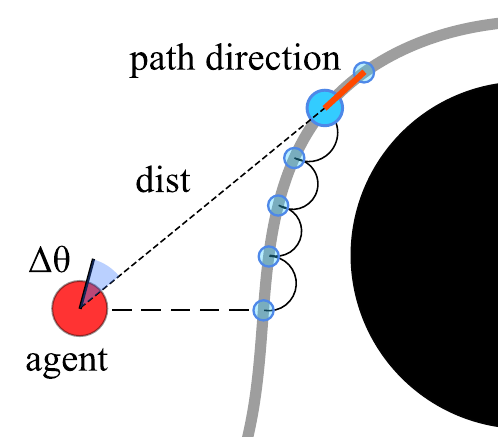}%
        \label{fig:obersvations-gp}%
    }
    \hfill
    \subfloat[]{%
        \includegraphics[width=0.30\linewidth]{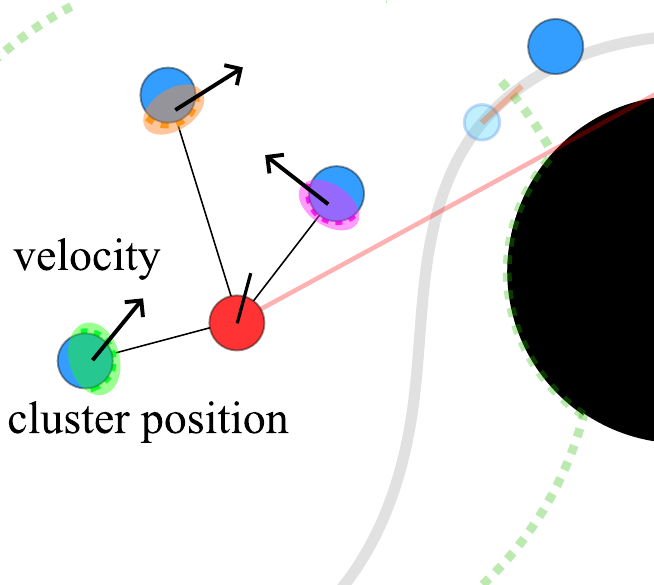}%
        \label{fig:obersvations-clusters}%
    }
    \caption{Overall illustration of the observation space. (a) Illustrates the environmental obstacles captured by the LiDAR, the velocity of the agent, and the goal distance. (b) Illustrates the global path observations, with the distance from the agent (red) to the target point (blue), the angle difference between the agent heading and the target point, and the direction of the global path at the target point. (c) Illustrates the estimated dynamic clusters extracted from the LiDAR. The position and velocity of each cluster are tracked over time and fed to the network as a graph.}
    \label{fig:obersvations}
\end{figure}

The observation space \( o_i^t \) is structured into five distinct components: \( o_i^t = [o^t_z, o^t_g, o^t_v, o^t_{gp}, o^t_C] \). The observations \(o_v^t, o_g^t, o_z^t\) are similar to those used in \cite{DRL-Nav} and are illustrated in Fig. \ref{fig:obersvations-lidar}. \(o_v^t\) represents the agent's current velocity and angular velocity, \(o_g^t\) is the local polar coordinates of the end goal, and \(o_z^t\) are the raw LiDAR measurements from the last three timesteps. The LiDAR used for our model is a 360-degree 2D LiDAR with 120 distance measurements, \(o^t_z \in \mathbb{R}^{3 \times 120}\), and a max range of 3.5 meters.

To effectively utilize global path information within our local navigation framework, we then introduce \( o_{gp}^t \), a running target point \(p_{\text{target}}\) that we represent in local polar coordinates along with the path's future direction. This approach ensures the robot maintains adherence to a pre-planned route while dynamically adjusting to its immediate environment.

\begin{equation}
  p_{\text{target}} = \text{GP}\left[\underset{i}{\text{argmin}}(|\text{GP}[i] - p_{\text{agent}}|) + \mathcal{H} \right],
\end{equation}

\noindent where \textit{GP} represents the given global path as a list of positions, \(p_{\text{target}}\) denotes the current position of the agent, and \(\mathcal{H}\) is the horizon, indicating how many points ahead of the nearest point we set as the target point.


To account for dynamic obstacles and facilitate collision avoidance, we introduce \(o_C\), which captures the positions and estimated velocities of neighboring dynamic clusters. This component leverages the LiDAR data to detect and track surrounding objects, ensuring the robot can navigate safely in dynamic environments.

\begin{equation}
o_C^t = \{ (p_{i}^t, v_{i}^t) \mid i \in \mathcal{N} \},
\end{equation}
where \( p_{i}^t \) denotes the position of the closest point in the \textit{i}-th neighboring dynamic cluster at time \textit{t}, measured in polar coordinates relative to the robot's current position, and \( v_{i}^t \) represents the estimated velocity of the \textit{i}-th cluster at time \textit{t}. The set \(\mathcal{N}\) includes all neighboring clusters detected within the LiDAR range.

Instead of providing the position of the cluster's center, we use the point within the cluster that is closest to the robot. This approach accounts for varying object radii by focusing on the nearest point, thereby allowing the network not to rely on a fixed or learned radius.

These clusters are extracted from the LiDAR measurements, and their trajectories are tracked over time using a model-free object detection and tracking approach introduced by~\cite{model_free_tracking}. This model-free method achieves dynamic object detection and tracking from raw LiDAR by clustering points based on motion patterns, requiring no object shape priors. A key aspect is that the velocity of neighboring dynamic clusters is estimated using a constant-velocity model, incorporating both linear and angular velocity components. This approach allows tracking without prior knowledge of object classes and assumes rigid-body motion for dynamic objects. Crucially, to distinguish between dynamic clusters and static obstacles with similar LiDAR measurements, the system employs a hierarchical data association framework. At a coarse level, clusters are assigned to static backgrounds or dynamic objects using an iterative closest point (ICP) algorithm with outlier rejection. At a finer level, associated clusters are further refined based on spatiotemporal consistency tests, ensuring accurate classification between static and dynamic objects. Although we use one particular method here, the approach is modular—any robust clustering and tracking method could be applied.

The resulting tracking information is represented as a graph, where nodes correspond to the clusters with node features such as relative distances to the closest point in the cluster and velocities, as illustrated in Fig. \ref{fig:obersvations-clusters}. This structured representation aids in efficient processing and interpretation of the dynamic environment, enabling the robot to perform cooperative maneuvers and avoid potential collisions effectively.

\subsection{Action space}
The action space of the differential robot is defined by its translational and rotational velocities, denoted as \(a_t = [v_t, w_t]\). Taking into account the kinematic constraints of the physical robot and practical applications, the translational velocity is limited to \(v \in (0.0, 1.0)\), while the rotational velocity is constrained to \(w \in (-1.0, 1.0)\). To ensure comparability with established benchmarks, we adopt the action space framework outlined in \cite{DRL-Nav, GA3C-CADRL}, which prohibits backward movement (i.e., \(v<0.0\)).

\begin{figure*}[!h]
    \centering
    \includegraphics[width=0.9\linewidth]{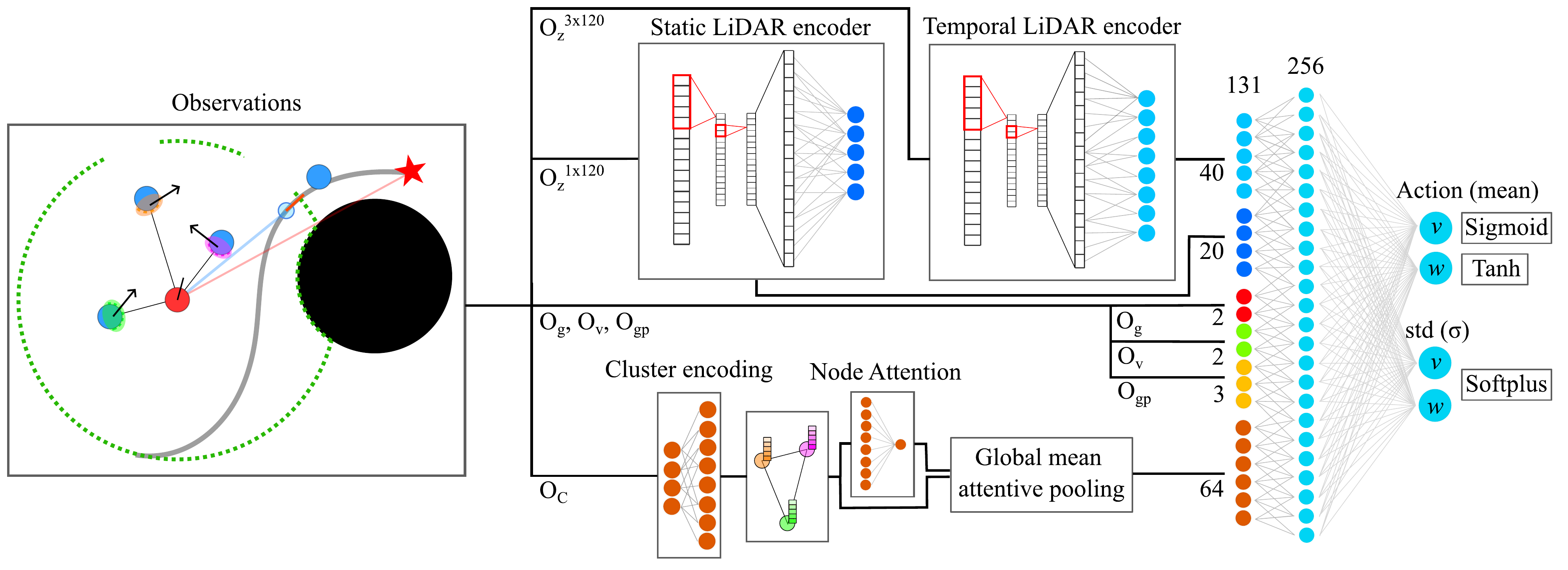}
    \caption{An illustration of the actor-network architecture. The network includes three observation encoders: a static LiDAR encoder that processes the current frame through Conv1D layers to capture static features, a temporal LiDAR encoder that processes the last three frames through Conv1D layers to learn dynamic features, and a neighbor attentive Graph Neural Network encoder. The combined encoded features are passed to fully connected layers and the network outputs the mean and standard deviation for the action normal distributions, which include linear and angular velocities. The critic network follows an identical structure but outputs state values instead.}
    \label{fig:network}
\end{figure*}
\subsection{Reward Function}
The reward function is designed to ensure that the agent navigates efficiently and safely in a crowded environment while adhering to the global path. It is composed of several components that balance goal achievement, collision avoidance, and social compliance.

The overall reward function is defined as follows:
\begin{equation}
r(s_i^t, a_i^t) =
\begin{cases} 
      r_{\text{goal}} & \text{if } s_t \in S_{\text{goal}} \\
      r_{\text{collision}} & \text{if } d_{\text{min}} < 0 \\
      r_{\text{social}} + r_{\text{progress}} & \text{otherwise}
\end{cases}
\end{equation}

In this reward structure, \( r_{\text{goal}} \) represents the reward given when the agent reaches the goal state \( S_{\text{goal}} \). This high reward encourages the agent to complete the navigation task successfully. The collision penalty, \( r_{\text{collision}} \), is applied if the agent collides with any obstacles or other agents, determined by \( d_{\text{min}} < 0 \). This penalty discourages collisions, promoting safer navigation.

The social distance penalty, \( r_{\text{social}} \), ensures the agent maintains an appropriate distance from other agents, respecting their personal space. It is calculated as:
\begin{equation}
r_{\text{social}} = r_s \left( \frac{PS - d_{\text{min}}}{PS} \right),
\end{equation}
where \( r_s \) is a penalty factor, \( PS \) is the personal space distance, and \( d_{\text{min}} \) is the closest distance to another agent within this personal space. This penalty mitigates the otherwise sparse collision penalty and facilitates quicker learning by maintaining a comfortable distance from other agents.

The progression reward, \( r_{\text{progress}} \), is designed to encourage the agent to make progress towards a dynamically moving target point along the global path. Traditional approaches calculate the progression reward based on the distance to the final goal, which may not adequately capture the agent's ability to follow a predefined path, especially in environments with dynamic obstacles. In our approach, the target point, denoted as \( p^{\text{target}}_t \), moves along the global path, ensuring the agent is continually incentivized to follow the planned route. The progression reward is computed as the difference between the distance from the agent's previous position \( p^{\text{agent}}_{t-1} \) to the target point and the distance from the agent's current position \( p^{\text{agent}}_t \) to the same running target point of timestamp $t$, expressed mathematically as:
\begin{equation}
r_{\text{progress}} = r_{\text{p}} \left( \lVert p^{\text{agent}}_{t-1} - p^{\text{target}}_t \rVert - \lVert p^{\text{agent}}_t - p^{\text{target}}_t \rVert \right),
\end{equation}
where \( r_{\text{p}} \) is the position shaping factor. This formulation ensures that the agent is rewarded for moving closer to the target point along the path, promoting adherence to the global path while retaining the flexibility to maneuver around dynamic obstacles.

\subsection{Model architecture}
The model architecture is designed to effectively handle the complex task of local navigation and collision avoidance in dynamic environments. The network leverages three key observation encoders to process diverse types of input data.
\medbreak
\textit{The static LiDAR encoder} focuses on capturing the static features of the environment. It processes the current LiDAR frame through a series of \ac{Conv1D} layers, which extract pertinent spatial information about stationary objects and obstacles.
\medbreak
\textit{The temporal LiDAR encoder} is responsible for learning dynamic features by analyzing how points move across time. It processes the last three LiDAR frames using \ac{Conv1D} layers, enabling the network to understand temporal changes and movements of objects, which is crucial for tracking and predicting dynamic obstacles.
\medbreak
\textit{The neighbor attentive Graph Neural Network encoder} serves a pivotal role in understanding interactions among dynamic clusters. It encodes the information about neighboring objects and their relationships using node attention mechanisms to highlight significant interactions and dependencies. Global mean attentive pooling is employed to ensure that the most relevant information is retained and emphasized.

\medbreak
After encoding the observations, the combined features from all three encoders along with \(o_g\), \(o_v\), and \(o_{\text{gp}}\) are passed to fully connected layers. The actor network then outputs the mean and standard deviation for the action normal distributions, specifically the linear and angular velocities required for navigation. This probabilistic approach enables the model to handle uncertainty and make more robust decisions in dynamic settings. The actor network is illustrated in Fig. \ref{fig:network}.

The critic network, which evaluates the value of each state, has a structure identical to that of the actor network. However, instead of outputting action distributions, it produces state values, which are essential for assessing the quality of the chosen actions.

\subsection{Training Setup}
\label{sec:training_setup}
A \ac{PPO} actor and critic is trained with the TorchRL data-driven framework\cite{bou2023torchrl}. In Table \ref{tab:hyperparameters} is shown the hyperparameters used to train the model.


\begin{table}[!ht]
        \centering
        \caption{Training hyperparameters}
        \begin{tabular}{cc ccc}
            \cmidrule{1-2}\cmidrule{4-5}
            Parameters & Values & & Parameters & Values\\ 
            \cmidrule{1-2}\cmidrule{4-5}
            learning rate critic & 4e-4 & & \(\Delta t\) & 0.1s \\
            learning rate actor & 2e-5 & &  \(\mathcal{H}\) & 5\\
            entropy coefficient & 0 & & \(r_p\) & 2.5\\
            reward discount & 0.99 & & \(r_s\) & -0.25m\\
            GAE discount & 0.95 & & PS & 0.3m \\
            PPO train epoch & 10 & & \(r_{\text{goal}}\) & 15\\
            clip value & 0.2 & &  \(r_{\text{collision}}\) & -25\\
            \cmidrule{1-2}\cmidrule{4-5}
        \end{tabular}
        \label{tab:hyperparameters}

\end{table}
The training environments are designed to provide diverse challenges to ensure robust policy learning and are created using the VMAS framework \cite{VMAS}. The environments are illustrated in Fig. \ref{fig:env_no_path_no_grid} and includes the following:

\begin{figure}[t]
   \centering
   \includegraphics[width=0.5\textwidth]{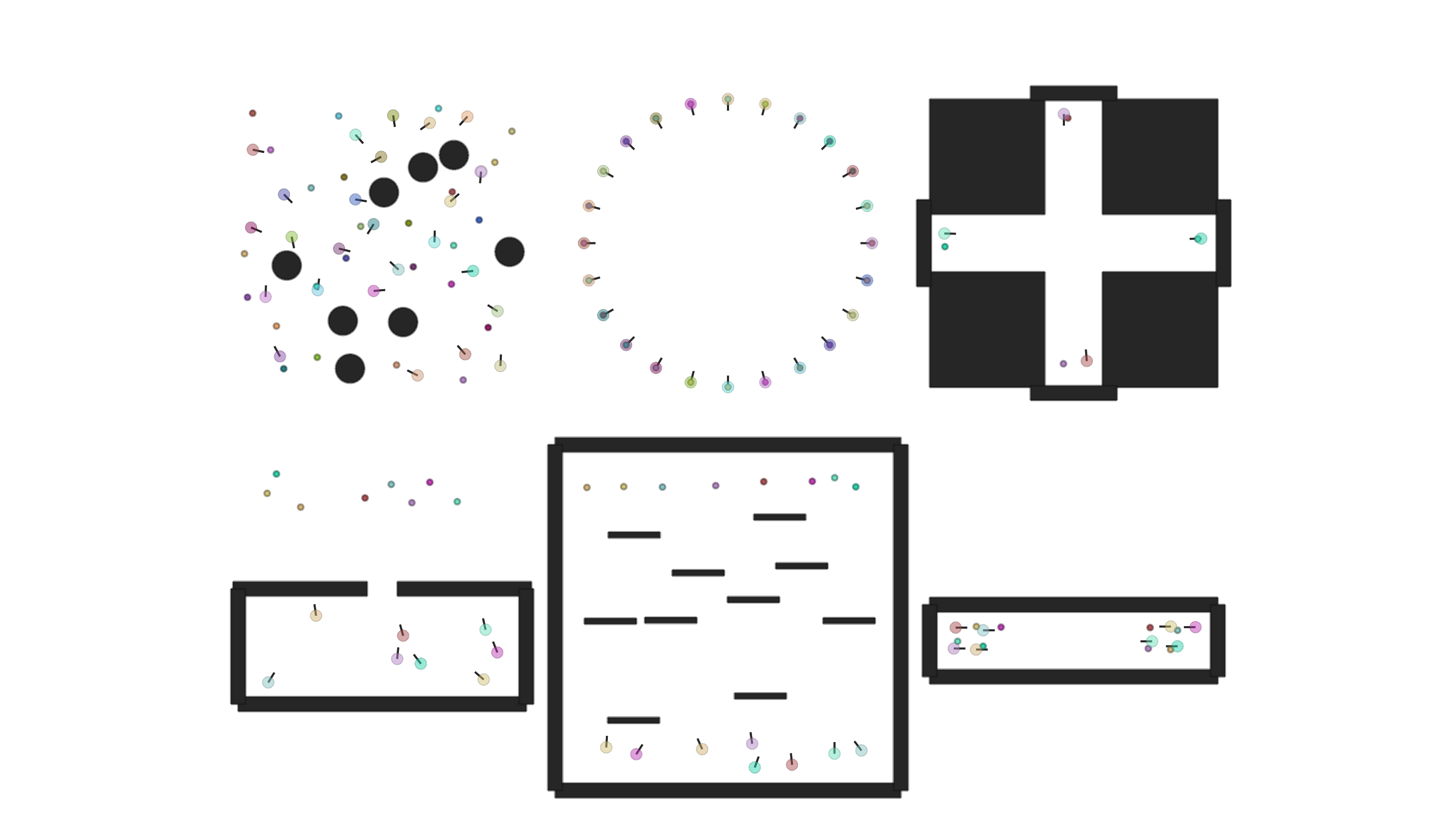}
   \caption{The six different environments used during training.}
   \label{fig:env_no_path_no_grid}
   \vspace{-4mm}
\end{figure}

Random Environment: In this dynamic scenario, obstacles, start positions, and goals are randomly generated within a 10x10 meter area. With 8 spherical obstacles and 25 agents, this environment fosters adaptability in navigation by simulating highly unpredictable conditions.

Circle Environment: Here, 24 agents are positioned along the circumference of a 10-meter diameter circle, with their goals directly opposite. This setup challenges the agents to navigate efficiently while avoiding collisions, as all agents converge toward their respective targets.

Plus Environment: This environment features two intersecting corridors, each measuring 2x10 meters, forming a plus sign. With 4 agents navigating these narrow passages, the scenario emphasizes precise maneuvering and effective collision avoidance in confined spaces.

Doorway Environment: In this scenario, agents must pass through a narrow doorway from a 10x4 meter room. The doorway, twice the diameter of the agents, allows only one to pass at a time. With 5 agents, this environment stresses coordination and timing to navigate the bottleneck without collisions.

Room Environment: This indoor scenario places 8 agents in a 10x10 meter area filled with 10 randomly positioned walls. Agents must navigate through these obstacles to reach their goals, simulating complex indoor environments that require careful pathfinding and obstacle avoidance.

Hallway Environment: Set within a 10-meter long, 2.5-meter wide corridor, this scenario requires 8 agents to swap positions while avoiding collisions. It tests the robots' ability to efficiently navigate and coordinate within a confined space, highlighting both individual and collective navigation skills.

%% file: sections/ablation_study.tex
\section{Ablation study}
\label{sec:ablation_study}
To validate the contribution of global path integration and \ac{GNN}-based agent interactions to our model's navigation performance, we conducted a targeted ablation study.  We evaluated two ablated variants against our full model to isolate the impact of each component: (a) Full Model, with both global path and \ac{GNN}; (b) No Global Path, ablating global path information, forcing reliance solely on the goal position; and (c) No \ac{GNN}, removing structured agent interaction via the \ac{GNN}.

We conducted ablation experiments in the doorway environment, a scenario deliberately chosen for its demanding nature due to high agent density, narrow passages requiring precise agent coordination, and complex obstacle layouts specifically designed to induce local minima, leading to potential agent entanglement and requiring long-horizon planning for efficient navigation.  This environment acutely tests both global path following and effective multi-agent coordination. Each model variant was evaluated across 250 trials in the doorway environment with 15 agents, measuring success rate, collision rate, and extra time.
\subsection{Results \& Discussion}
\begin{table}[b]
    \centering
    \caption{Ablation Study Results in the Doorway Environment \\with 15 Agents.}
    \resizebox{\columnwidth}{!}{%
    \begin{tabular}{lccc}
        \toprule
        \textbf{Metric} & \textbf{Full Model } & \textbf{No Global Path} & \textbf{No GNN} \\
        \midrule
        Success Rate (\%)             & \textbf{96.7} & 92.7 & 93.2 \\
        Stuck / Collision Rate (\%)   & \textbf{3.3 / 0.} & 7.1 / 0.2 & 2.2 / 4.6 \\
        Extra Time (Normalized)       & 18.7 & 21.8 & \textbf{14.7} \\
        Average Speed (m/s)           & 0.42  & 0.34 & \textbf{0.57} \\
        \bottomrule
    \end{tabular}%
    }
    \label{tab:ablation_results}

\end{table}
The ablation results are presented in Table \ref{tab:ablation_results}, clearly highlighting the critical role of both components. Ablating the global path information significantly impaired navigation: the No Global Path variant exhibited substantially increased extra time and decreased success rates, demonstrating its struggle with efficient long-range navigation and goal attainment within the confines of the doorway. This performance degradation is mainly due to agents becoming congested near the doorway and exhibiting an excessive focus on reaching intermediate goal locations, leading to navigation inefficiencies. 
Similarly, ablating \ac{GNN} interactions led to degraded collision avoidance: the No \ac{GNN} variant suffered an increase in collision rate within the dense multi-agent doorway environment, indicating the \ac{GNN}'s crucial role in mediating safe and effective agent interactions. Without the GNN's structured interaction modeling, the No GNN variant becomes less adept at anticipating agent movements and resolving potential conflicts, resulting in a higher frequency of collisions.

These ablation results clearly confirm the synergistic benefits of integrating both global path guidance and \ac{GNN}-based agent interaction in our navigation model. The performance degradation observed when ablating either component in a challenging environment unequivocally validates our core architectural choices, demonstrating their combined necessity for robust and efficient multi-robot collision avoidance.


%% file: sections/experiments.tex
\section{Experimental setup}
\label{sec:experiments}
\subsection{Evaluation criteria}
To assess the performance of our model, we use four main criteria.
\begin{itemize}
    \item \textit{Success rate:} is the ratio of the number of robots reaching their goals within a certain time limit without any collisions over the total number of robots.
    \item \textit{Collision/stuck rate:} is the ratio of the number of robots that collide or do not reach their goals within two minutes over the total number of robots.
    \item \textit{Extra time:} measures the difference between the travel time averaged over all robots and the lower bound of the travel time along the path (i.e. the average cost time of going along the path toward the goal for robots at max speed).
    \item \textit{Average speed:} measures the average speed of the robot team during navigation.
\end{itemize}
\input{LargeObjects/ResultsTable.tex}

\subsection{Baselines}
We compare our model against three different baselines: NH-ORCA, DRL-NAV, and GA3C-CADRL. NH-ORCA~\cite{NH-ORCA} is a heuristic-based approach that employs optimal reciprocal collision avoidance techniques to guide robots in dynamic environments. DRL-NAV~\cite{DRL-Nav} is a deep reinforcement learning-based navigation approach that we retrained according to the original paper, as the released model showed poor results and lacked normalization of the observations. GA3C-CADRL~\cite{GA3C-CADRL} is an agent-based model that uses processed information about agents in the environment rather than raw sensor input. Since this model supports only sphere-structured observations, we modeled the environment as a series of spheres to ensure compatibility and allow the model to have a higher angular velocity, up to \(5.23 \text{ rad/s}\), to match the trained action outputs.

To evaluate model robustness under realistic sensor imperfections, we incorporated sensor noise into the observation inputs for all experiments. Mimicking typical sensor inaccuracies, we applied noise to agent positional information \(\pm 0.1m\), velocity information \(\pm 0.1m/s\), and LiDAR measurements (\(\pm 3.5\%\) Gaussian noise).


Following the approach in \cite{GP_as_goal}, we assume that none of these methods are designed to function as standalone global navigation algorithms. They lack global information, which would result in encountering many local minima. Therefore, the observation of the goal position for all baseline models is set to the running target point on the global path. Each agent's global path is computed independently using $A^*$, an algorithm that provides optimal paths in static environments, without considering other agents.
\subsection{Evaluation Scenarios}
The evaluation scenarios chosen are; Circle, Doorway, Hallway, and Random as described in Section \ref{sec:training_setup}.  For testing, each environment is scaled to 15 meters in size, allowing for the deployment of varying numbers of agents to assess the robustness and scalability of the trained policy.

%% file: LargeObjects/ResultsTable.tex
\begin{table*}[!ht]
\centering
\rowcolors{3}{}{white}
\caption{Model performance on evaluation scenarios}
\label{table:result_comparison}
\resizebox{\textwidth}{!}{\begin{tabular}{| c | c |c c c|c c c|c c c|c c c|}
\hline
\multicolumn{2}{|c|}{Training settings} & \multicolumn{12}{c|}{Test environments with variable number of agents: } \\
\hline
\multirow{2}{*}{Metrics}  & \multirow{2}{*}{Methods}
& \multicolumn{3}{c|}{Circle (radius 7.5m)}
& \multicolumn{3}{c|}{Doorway} 
& \multicolumn{3}{c|}{Narrow Corridor}
& \multicolumn{3}{c|}{Random (8 obstacles)} \\ 
                                &
                                &  10 & 20 & 40 
                                &  5 & 10 & 15 
                                &  8 & 12 & 16 
                                &  10 & 20 & 40\\
\hline\hline
\rowcolor{lightgray}\cellcolor{white}               
                                & \cellcolor{white} NH-ORCA \cite{NH-ORCA}
                                & \textbf{100.} & \textbf{100.} & \textit{72.5}    
                                & \textit{32.} & \textit{31.} & \textit{32.7}   
                                & \textit{95.} & \textit{91.67} & \textit{80.}   
                                & \textit{93} & \textit{95.5} & \textit{97.25} \\ 

\cellcolor{white}
                                & DRL-Nav\cite{DRL-Nav}
                                & \textbf{100.} & \textbf{100.} & \textit{5.}   
                                & \textbf{100.} & \textbf{100.} & \textit{91.3}   
                                & \textit{93.8} & \textit{90.} & \textit{82.5}   
                                & \textit{99.} & \textit{97.} & \textit{95.3}\\ 
                                
\rowcolor{lightgray}\cellcolor{white}                                
                                & \cellcolor{white}GA3C-CADRL \cite{GA3C-CADRL}
                                & \textbf{100.} & \textit{60.} & \textit{52.5}   
                                & \textit{28.} & \textit{20.} & \textit{14.7}   
                                & \textit{73.75} & \textit{53.17} & \textit{20.}   
                                & \textbf{100.} & \textit{95.} & \textit{83.8}\\ 
                                
\cellcolor{white}
\multirow{-4}{*}{Success Rate (\%)}
                                & \cellcolor{white}  GIANT [ours]
                                & \textbf{100.} & \textbf{100.} & \textbf{95.}   
                                & \textbf{100.} & \textit{99.} & \textbf{96.67}   
                                & \textbf{100.} & \textbf{100.} & \textbf{97.5}  
                                & \textbf{100.} & \textbf{99.5} & \textbf{99.25}\\ 

\hline \hline
\rowcolor{lightgray}\cellcolor{white}               
                                & \cellcolor{white} NH-ORCA \cite{NH-ORCA}
                                & \textbf{0. / 0.} & \textbf{0. / 0.} & \textit{10. / 17.5}    
                                & \textit{4. / 64.} & \textit{19. / 50.} & \textit{22. / 45.33}   
                                & \textit{0 / 13.75} & \textit{2.5 / 14.17} & \textit{6.25 / 19.37}   
                                & \textit{0. / 7.} & \textit{0. / 4.5} & \textit{0. / 2.75}\\

\cellcolor{white}
                                & DRL-Nav\cite{DRL-Nav}
                                & \textbf{0. / 0.} & \textbf{0. / 0.} & \textit{15. / 80.}   
                                & \textbf{0. / 0.} & \textbf{0. / 0.} & \textit{6. / 2.7}   
                                & \textit{5. / 0.} & \textit{8.33 / 0.} & \textit{20. / 0.}   
                                & \textit{1. / 0.} & \textit{3. / 0.} & \textit{4.2 / 0.5}\\
                                
\rowcolor{lightgray}\cellcolor{white}                                
                                & \cellcolor{white}GA3C-CADRL \cite{GA3C-CADRL}
                                & \textbf{0. / 0.} & \textit{0. / 40.} & \textit{0. / 47.5}   
                                & \textit{4. / 68.} & \textit{2. / 78.} & \textit{0.6 / 84.6}   
                                & \textit{0. / 26.25} & \textit{0. / 45.83} & \textit{0. / 80.}   
                                & \textbf{0. / 0.} & \textit{0. / 5.} & \textit{0.5 / 15.7}\\
                                
\cellcolor{white}
\multirow{-4}{*}{Stuck/Collision Rate (\%)}
                                & \cellcolor{white}  GIANT [ours]
                                & \textbf{0. / 0.} & \textbf{0. / 0.} & \textbf{0. / 5.}   
                                & \textbf{0. / 0.} & \textit{1. / 0.} & \textbf{3.3 / 0.}   
                                & \textbf{0. / 0.} & \textbf{0. / 0.} & \textbf{2.5 / 0.}  
                                & \textbf{0. / 0.} & \textbf{0.5 / 0.} & \textbf{0.75 / 0.} \\
                                
\hline \hline
\rowcolor{lightgray}\cellcolor{white}               
                                & \cellcolor{white} NH-ORCA \cite{NH-ORCA}
                                & \textit{6.9} & \textit{15.34} & \textit{56.16}   
                                & \textit{75.} & \textit{77.39} & \textit{78.89}   
                                & \textit{23.24} & \textit{28.69} & \textit{43.06}   
                                & \textit{10.30} & \textit{11.10} & \textit{14.07}\\

\cellcolor{white}
                                & DRL-Nav\cite{DRL-Nav}
                                & \textit{5.24} & \textbf{8.45} & \textit{100.37}   
                                & \textbf{2.21} & \textbf{5.54} & \textbf{16.51}   
                                & \textit{8.1} & \textit{14.8} & \textit{29.9}   
                                & \textit{2.34} & \textit{5.12} & \textit{9.09}\\
                                
\rowcolor{lightgray}\cellcolor{white}                                
                                & \cellcolor{white}GA3C-CADRL \cite{GA3C-CADRL}
                                & \textbf{5.23} & \textit{46.86} & \textit{56.76}   
                                & \textit{79.02} & \textit{87.80} & \textit{93.42}   
                                & \textit{31.42} & \textit{53.35} & \textit{88.74}   
                                & \textit{3.36}  & \textit{9.16} & \textit{21.8}\\
                                
\cellcolor{white}
\multirow{-4}{*}{Extra Time (ratio)}
                                & \cellcolor{white}  GIANT [ours]
                                & \textit{5.92} & \textit{8.69} & \textbf{23.89}   
                                & \textit{4.38} & \textit{10.9} & \textit{18.71}   
                                & \textbf{4.02} & \textbf{14.18} & \textbf{20.92}   
                                & \textbf{1.54} & \textbf{3.15} & \textbf{6.59}\\
                                
\hline \hline
\rowcolor{lightgray}\cellcolor{white}               
                                & \cellcolor{white} NH-ORCA \cite{NH-ORCA}
                                & \textit{0.74} & \textit{0.57} & \textit{0.25}   
                                & \textit{0.09} & \textit{0.1} & \textit{0.1}   
                                & \textit{0.43} & \textit{0.28} & \textit{0.22}   
                                & \textit{0.7} & \textit{0.57} & \textit{0.44}\\
                                
\cellcolor{white}
                                & DRL-Nav\cite{DRL-Nav}
                                & \textit{0.82} & \textbf{0.73} & \textit{0.06}   
                                & \textbf{0.87} & \textbf{0.76} & \textbf{0.50}   
                                & \textit{0.71} & \textbf{0.47} & \textit{0.32}   
                                & \textit{0.87} & \textit{0.69} & \textit{0.58}\\
\rowcolor{lightgray}\cellcolor{white}                                
                                & \cellcolor{white}GA3C-CADRL \cite{GA3C-CADRL}
                                & \textbf{0.84} & \textit{0.56} & \textit{0.44}   
                                & \textit{0.11} & \textit{0.09} & \textit{0.07}   
                                & \textit{0.64} & \textit{0.18} & \textit{0.14}   
                                & \textit{0.81} & \textit{0.72} & \textit{0.31}\\
                                
\cellcolor{white}
\multirow{-4}{*}{Average Speed (m/s)}
                                & \cellcolor{white}  GIANT [ours]
                                & \textit{0.79} & \textit{0.71} & \textbf{0.48}   
                                & \textit{0.76} & \textit{0.55} & \textit{0.42}   
                                & \textbf{0.78} & \textit{0.46} & \textbf{0.36}   
                                & \textbf{0.88} & \textbf{0.77} & \textbf{0.62}\\

\hline
\end{tabular}}
\vspace{-4mm}
\end{table*}

%% file: sections/results.tex
\section{Results}
\label{sec:results}

The results in Table \ref{table:result_comparison} demonstrate the robust performance of our model across a variety of challenging evaluation scenarios in comparison to the baseline methods NH-ORCA, DRL-NAV, and GA3C-CADRL. A key factor in our model's success is its ability to effectively differentiate between the global path and the final goal. Our approach discerns between temporary waypoints and final objectives, unlike baseline models that treat the goal as a running target along the global path. This capability allows the model to balance following the global path and reaching the final destination, consistently achieving higher success rates across all scenarios.

Inferring agent states from LiDAR and tracking introduces inherent observation noise, enhancing the model's robustness in dynamic environments. This robustness is evident in our model's lower collision and stuck rates, particularly in complex scenarios like the Narrow Corridor and Doorway. Effectively navigating with noisy observations ensures safer and more reliable real-world operation with imperfect sensor data.

In terms of navigation efficiency, the DRL-NAV model exhibits quicker completion times in certain scenarios, such as the Circle and Doorway. However, this speed is often accompanied by higher collision rates, which compromises overall reliability. Our model, on the other hand, consistently records lower extra time ratios, demonstrating a more balanced approach between speed and safety. For instance, in the Doorway scenario with 15 agents, our model's extra time of 18.71 is comparable to DRL-NAV's 16.51, yet it maintains a higher success rate and lower collision rate, indicating a better trade-off between efficiency and safety.

NH-ORCA and DRL-NAV show strong performance in some scenarios, particularly with smaller agent counts, but struggle in more crowded environments, such as the Circle with 40 agents and the Narrow Corridor, where they record higher stuck/collision rates. This highlights our model's advantages in maintaining high performance across varying agent densities and environmental complexities.

Finally, a significant strength of our model is its adaptability across diverse environments, as shown by its consistent performance in the Doorway, Hallway, and Random scenarios. This adaptability indicates that our model generalizes effectively across different environmental structures and agent densities, which is crucial for robust performance in multi-robot systems. In contrast, GA3C-CADRL, which was not trained in confined or narrow spaces, shows significantly lower performance in the Doorway and Hallway scenarios, underscoring its limitations in handling environments with tight spatial constraints. The consistently high success rates and low collision rates across all tested environments emphasize our approach's robustness in managing complex and unpredictable scenarios.


%% file: sections/conclusion.tex
\section{Conclusion and future work}
\label{sec:conclusion}
In this paper, we presented a novel multi-robot collision avoidance approach integrating global path planning with local navigation via attentive graph neural networks. Our method leverages pre-planned global paths to enable robots to maintain adherence to optimal routes while dynamically adapting to environmental changes. The evaluation of our model against established baselines; NH-ORCA, DRL-NAV, and GA3C-CADRL, across challenging scenarios with sensor noise, demonstrated its superior performance. Our model consistently achieved higher success rates and lower collision rates, especially in complex environments where baselines struggled.  The ability to differentiate between global paths and final goals was crucial to our model's success, enabling more effective balancing of path adherence with goal achievement compared to existing methods.

This research suggests several promising directions to further enhance our multi-robot collision avoidance method's capabilities and generalizability. First, generalizing to heterogeneous agents is crucial for real-world use. Beyond homogeneous settings, future models must handle diverse robots. Investigating robustness to non-cooperative, adversarial agents is warranted, requiring diverse training and explicit resilience evaluations. Second, we will explore cooperative reward mechanisms to enhance multi-agent performance. Our current single-agent reward solution, a pragmatic choice for decentralized, non-communicating robots, can be advanced by investigating direct cooperative rewards without explicit communication. Game-theoretic approaches offer a way to predict agent cooperativeness implicitly. Leveraging game theory to model interactions and predict cooperative levels will inform the design of implicit cooperative rewards, fostering collaboration beyond individual objectives.